%% file: _final_paper.tex
\documentclass{article}

\usepackage[nolist]{acronym}
\input{acronyms}

\usepackage{listings}

\usepackage{microtype}
\usepackage{graphicx}
\usepackage{subfigure}
\usepackage{booktabs} 
\usepackage{multirow}

\usepackage[T1]{fontenc}
\usepackage[utf8]{inputenc}

\usepackage{hyperref}

\newcommand{\Return}{\State \textbf{return} }

\usepackage[accepted]{icml2025}

\usepackage{amsmath}
\usepackage{amssymb}
\usepackage{mathtools}
\usepackage{amsthm}
\usepackage{svg}
\usepackage[capitalize,noabbrev]{cleveref}
\usepackage[linesnumbered,ruled,vlined]{algorithm2e}

\theoremstyle{plain}

\theoremstyle{definition}

\theoremstyle{remark}

\usepackage[textsize=tiny]{todonotes}

\icmltitlerunning{Do Vision-Language-Models show human-like logical problem-solving capability in point and click puzzle games?}

\begin{document}

\twocolumn[
\icmltitle{Do Vision-Language-Models show human-like logical problem-solving capability in point and click puzzle games?}



\icmlsetsymbol{equal}{*}

\begin{icmlauthorlist}
\icmlauthor{Maximilian Triebel}{equal,yyy}
\icmlauthor{Marco Menner}{equal,yyy}
\icmlauthor{Dominik Helfenstein}{equal,yyy}
\end{icmlauthorlist}

\icmlaffiliation{yyy}{Institute of Artificial Intelligence, University of Stuttgart, Stuttgart, Germany}

\icmlcorrespondingauthor{}{}

\icmlkeywords{Machine Learning, Vision Language Models, Benchmark}

\vskip 0.3in
]



\printAffiliationsAndNotice{\icmlEqualContribution} 


\input{abstract}

\section{Introduction}
\label{sec:introduction}
\input{introduction}

\section{Related Work}
\label{sec:related-work}
\input{related_work}


\section{The VLATIM Benchmark}
\label{sec:the-vlatim-bench}
\input{the_vlatim_benchmark}

\section{Experiments}
\label{sec:experiments}
\input{experiments}

\section{Conclusion}
\label{sec:summary}
\input{summary}

\section{Outlook}
\label{sec:outlook}
\input{outlook}




\bibliography{references}
\bibliographystyle{icml2025}

\newpage
\appendix
\onecolumn

\section{Metrics}
\label{appendix:metrics}

The \acp{bbox} are scored using \ac{IoU} between ground truth and prediction, as well as euclidean distance between their centers, as an \ac{IoU} of $0$ provides no information of whether the \ac{bbox} is far away or close. 
\acp{bbox} are used in Part 1 for object localization and in Part 3 for predicted target regions.

For classification and single-choice tasks, the success rate is being measured via string matching, which rely on the models to follow the prompts and output a single given choice and nothing else. 
These are used in Part 1 for object classification, Part 2 for single-choice. For Part 4 and 5, human evaluation is necessary to mark each run as success or failure, and here the success rate is being taken as well.

For the textual component of Part 3, we utilized an LLM-based evaluation framework. Unlike Part 2, the responses here required semantic analysis. We provided an evaluator LLM namely Gemini 2.5 Flash with a prompt to verify whether the candidate models output was semantically similar and logically consistent with the ground truth. The evaluator was constrained to output a binary ``True'' or ``False''. This methodology allowed for a scalable and efficient examination of model performance.

\section{Framework}
\label{appendix:framework}
\input{appendix_framework}

\newpage

\section{The Incredible Machine 2}
\label{appendix:tim}
\begin{figure}[h]
    \centering
    \includegraphics[width=0.7\linewidth]{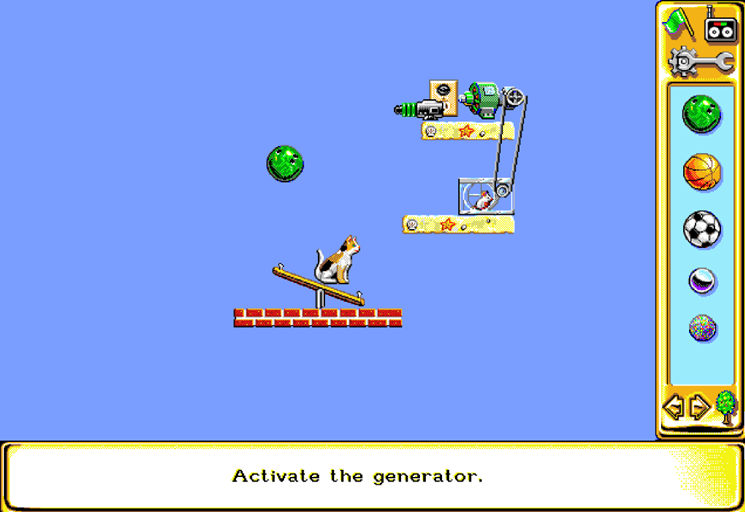}
    \caption{\ac{TIM} game interface with playfield in the middle (blue area), parts bin as well as navigation buttons on the right and task description at the bottom.}
    \label{fig:placeholder}
\end{figure}

\section{Action Space}
\label{appendix:tasks}
\input{appendix_tasks}

\section{Prompts}
\label{appendix:prompts}
\input{appendix_prompts}



\end{document}

%% file: acronyms.tex
\acrodef{AI}{Artificial Intelligence}
\acrodef{BALROG}{Benchmarking Agentic LLM/VLM Reasoning On Games}
\acrodef{Gemini}{Gemini 2.5 Flash}
\acrodef{GPT}{GPT-5 mini}
\acrodef{GUI}{graphical user interface}
\acrodef{HUD}{Heads-Up Display}
\acrodef{IoU}{Intersection over Union}
\acrodef{iVISPAR}{Interactive Visual-Spatial Reasoning}
\acrodef{JSON}{JavaScript Object Notation}
\acrodef{OOD}{out-of-districution}
\acrodef{Qwen2.5}{Qwen2.5-VL-7B-Instruct}
\acrodef{Qwen3}{Qwen3-VL-235B-A22B-Instruct}
\acrodef{QA}[Q\&A]{Question and Answer}
\acrodef{RL}{reinforcement learning}
\acrodef{bbox}{Bounding box}
\acrodefplural{bbox}[bboxes]{bounding boxes}
\acrodef{SFT}{Supervised Fine-Tuning}
\acrodef{SOTA}{state-of-the-art}
\acrodef{TIM}{The Incredible Machine 2}
\acrodef{UI-Tars}{UI-Tars-1.5-7B}
\acrodef{VLA}{Vision-Language-Action Model}
\acrodefplural{VLA}[VLAs]{Vision-Language-Action Models}
\acrodef{VGRP}{Visual Grid Reasoning Puzzle}
\acrodef{VLATIM}{Vision-Language Against The Incredible Machine}
\acrodef{VLM}{Vision-Language(-Action) Model}
\acrodefplural{VLM}[VLMs]{Vision-Language(-Action) Models}

%% file: abstract.tex
\begin{abstract}
\acp{VLM} are increasingly applied to interactive environments, yet existing benchmarks often overlook the complex physical reasoning required for point-and-click puzzle games. 
This paper introduces \ac{VLATIM}, a benchmark designed to evaluate human-like logical problem-solving capabilities within the classic physics puzzle game \ac{TIM}. Unlike existing benchmarks, \ac{VLATIM} specifically targets the critical gap between high-level logical reasoning and continuous action spaces requiring precise mouse interactions. This benchmark is structured into five progressive parts, assessing capabilities that range from basic visual grounding and domain understanding to multi-step manipulation and full puzzle solving. Our results reveal a significant disparity between reasoning and execution. While large proprietary models demonstrate superior planning abilities, they struggle with precise visual grounding. Consequently, they do not yet show human-like problem-solving capabilities.
\end{abstract}

%% file: introduction.tex
\acp{VLA} have gained increasing popularity in recent years as a promising approach to integrating visual perception, language understanding, and action generation within a unified framework. By bridging these modalities, \acp{VLA} enable agents to interpret complex visual scenes, follow natural language instructions, and perform contextually appropriate actions. Recent explorations have applied \acp{VLA} to open-world environments and interactive computer games such as \textit{Minecraft} \cite{li_jarvis-vla_2025}.
It remains an open question whether specialized \ac{VLA} architectures are strictly necessary for such tasks. Emerging research challenges the need for custom action heads, demonstrating that a simpler strategy, prompting a \ac{VLM} to output actions as text strings, is surprisingly powerful \cite{goyal_vla-0_2025}. This insight suggests that the reasoning capabilities of standard \acp{VLM} are sufficient for acting in complex environments, provided the action space can be textualized. This realization strongly supports the recent shift towards utilizing language-driven agents for autonomous interaction and problem-solving \cite{huo2026abotclawfoundationpersistentcooperative}.
In this context, a promising, yet underexplored area for \ac{VLM} research is point-and-click puzzle games, which require reasoning about physical mechanisms, cause-and-effect relationships, and creative problem-solving. Amongst others, \ac{TIM}, a classic puzzle series from the early 1990s known for its Rube Goldberg-style contraptions, stands out as an interesting environment for examining how \acp{VLM} can understand and manipulate complex, physics-based contraptions to achieve simple goals. With \ac{VLATIM}, we introduce a well defined benchmark exploring the question whether \acp{VLM} show human-like logical problem-solving capability in point-and-click puzzle games. Our benchmark tests \acp{VLM} on a wide range of human-like capabilities such as scene understanding, spatial reasoning, multi-step planning and action execution quantitatively as well as qualitatively pointing out their strengths and weaknesses. Ultimately, our findings provide a definitive answer to our core research question: current models fall significantly short of human-like problem-solving in these environments.

%% file: related_work.tex
The recent advancements in \ac{VLM} research have shifted evaluation benchmarks from static image analysis to dynamic, interactive environments, focusing on capabilities like long-horizon planning and visual spatial reasoning.  \textbf{\ac{BALROG}} \cite{paglieri_balrog_2025} for example assesses the agentic capabilities of \acp{VLM} on long-horizon tasks through a diverse set of challenging games, testing their ability to plan, reason spatially, and explore in dynamic environments. They employ detailed runtime feedback and prompting strategies. Their benchmark reveals that while current models show some success on simpler tasks, they struggle with more complex, procedurally generated environments, especially when vision-based decision-making is involved. \textbf{VideoGameBench} \cite{zhang_videogamebench_2025} even goes one step further, evaluating \ac{VLM}-based agents on full game completion across 23 titles using only raw visual inputs and core objectives like defeating the game's final boss. They observe that while the models can perform basic interactions such as movement, menu navigation, and simple combat, they consistently struggle with higher-order cognitive tasks including strategic planning, spatial reasoning, objective maintenance, and adaptive problem-solving. In the domain of logic puzzles, \textbf{\ac{VGRP}-Bench} \cite{ren_vgrp-bench_2025} focuses on visual grid reasoning (e.g. Sudoku), assessing perception and rule-following, while the \textbf{\ac{iVISPAR} Benchmark} \cite{mayer_ivispar_2025} utilizes sliding tile puzzles to provide rigorous metrics on spatial reasoning and alignment as well as deviation from optimal paths. Again, both benchmarks reveal that \acp{VLM} struggle with more complex configurations and problem properties.

Despite these contributions, we identify some limitations in existing methodologies. Action spaces are often either too simple and restricted to discrete inputs as in \textbf{VGRP-Bench} and \textbf{\ac{iVISPAR}} or basic keyboard movement as in \textbf{\ac{BALROG}}, or they are excessively complex, such as open-world games (e.g. \textit{Zelda}) \cite{zhang_videogamebench_2025}. There is a lack of benchmarks that focus on balancing logical thinking with spatial reasoning within a continuous action space, such as precise mouse interactions. Furthermore, prompting strategies often lack fairness. \textbf{VideoGameBench} withholds user manual information available to humans by only providing a high-level description of objectives and controls, resulting in near 0\% success rates that obscure the root causes of failure, whereas \textbf{\ac{BALROG}} provides excessive state feedback that artificially simplifies the task. Our work aims to bridge these gaps by providing a balanced environment with continuous action spaces and realistic information availability.

%% file: the_vlatim_benchmark.tex
\subsection{The Incredible Machine 2} 
We selected the point and click puzzle game \ac{TIM} as our benchmark environment. It is known for its over-complicated chain reactions designed to solve simple tasks. For instance, instead of simply flipping a switch to turn on an electric mixer, the player must solve a puzzle in which a bowling ball falls onto a teeter totter which launches a cat into the air towards a mouse stuck in a flywheel causing it to run away from the cat and therefore powering a generator which supplies the mixer with electricity (see appendix \ref{appendix:tim}).
\ac{TIM} is an appealing environment for our benchmark as it requires advanced reasoning to understand the scene and to come up with a solution, turn the solution into a sequence of actions and finally perform these actions with the mouse.

\subsection{Benchmark structure}

The benchmark is divided into five parts, ranging from easiest to hardest. The idea is to start with a minimal task and expand the scope of each part until all aspects of the game are encompassed in the final part. The first three parts are based on screenshots of the game and can be solved without a game interface. Only the last two parts require the models to interact with the game.

By providing five different parts, we intend to cover a diverse set of perception and reasoning tasks. The first part, ``visual grounding'', covers basic object classification and localization via \acp{bbox}. The second part, ``domain understanding'', evaluates text answers about different physical properties of objects and small combinations of them within the game. The third part, ``event reasoning'', requires the models to reason about cause, effect and outcomes of small puzzles with both textual and visual output. Part four, ``manipulation'', tests the models' ability to place, move, rotate and further interact with objects in the game. Part five, ``full puzzle solving'', extends part four by providing solely the task to solve the puzzle, the actual objective of the puzzle has to be read from the screen.

Table \ref{tab:overview} provides an overview of the benchmark's structure including information about the prompt, method, metrics and puzzle count used in each part. Details about prompts can be found in appendix \ref{appendix:prompts}. The five parts are described in section \ref{sec:part1_grounding} to \ref{sec:part5_full_puzzle}.

\begin{table*}[!ht]
    \centering
    \footnotesize
    \setlength{\tabcolsep}{3pt} 
    \renewcommand{\arraystretch}{1.2}
    \begin{tabular}{|l|p{6cm}|l|l|l|l|}
        \hline
        \textbf{Part} & \textbf{Task} & \textbf{Prompt} & \textbf{Method} & \textbf{Metrics} & \textbf{Count} \\ \hline
        \multirow{2}{*}{Visual Grounding} & Classify & PL & SC & SR & 40 \\ \cline{2-6} 
         & Localize Single / Multi & PL & bbox & IoU, ED & 40 / 40 \\ \hline
        \multirow{2}{*}{Domain Understanding} & Property Identification & PL / PL,PD & SC & SR & 20 / 20\\ \cline{2-6} 
         & State identification & PL & SC & SR & 20 \\ \hline
        \multirow{2}{*}{Event Reasoning} & Textual
         & - / PL / PD & LLM-E & SR & 10 per category \\ \cline{2-6} 
         & Visual & - / PL / PD & bbox & IoU, ED & 10 per category \\ \hline
        Manipulation & Move, Place, Rotate, Multi, Remove, Stretch & GI, AI & H-E & SR & 6 per category \\ \hline
        Full Puzzle Solving & Solve the puzzle & PD, GI, AI & H-E & SR & 6 \\ \hline
    \end{tabular}
    \caption{Benchmark structure overview: PL (Parts List), PD (Parts Descriptions), GI (Game Instructions), AI (Actions Instructions), SC (Single Choice), \ac{bbox}, LLM-E (LLM-Eval), H-E (Human-Eval), SR (Success Rate), \ac{IoU}, ED (Euclidean Distance)}
    \label{tab:overview}
\end{table*}

\subsection{Puzzle Design}
We apply no restrictions to the pool of objects selectable for puzzle design. All objects which are part of the puzzle are always mentioned in either the prompt
or the in-game task description. For the fourth and fifth part
of the benchmark, all puzzles are designed in such a way that they are solvable in one go, i.e. they do not require iterative trial and error by nature as this is also not within the scope of our research question. To give an illustrative example, one could think of having to place a cannon such that its ball shoots directly into a fixed pipe facing upwards which would require perfect estimation of the ball's trajectory. Furthermore, we vary the amount of fixed and movable solution-relevant and -irrelevant elements in the parts bin and on the playfield. We also make use of scenery further altering the degree of potential confusion or misidentification risks via e.g. texture overlaps with the only restriction to not modify the playfield's actual background. By manually specifying the puzzle games, we ensure the benchmark features a wide variety of different objects, physical mechanics, and task difficulties.

\subsection{Part 1: Visual Grounding}
\label{sec:part1_grounding}

Visual grounding is divided into object classification as well as localization of single or multiple objects, which allows us to test the model's zero-shot capabilities using single choice \ac{QA} and \acp{bbox}. As an \ac{IoU} score of $0$ does not contain any information about the positional offset of the predicted \ac{bbox}, we additionally provide the Euclidean distance between \acp{bbox} centers. A representative task involves the model localizing a candle within the scene.

This part forms the foundation of our benchmark. It helps identify the root causes of failure at the lowest level of perceptual ability required for self-orientation within the game.

\subsection{Part 2: Domain Understanding}
\label{sec:part2_understanding}

Domain understanding evaluates the models' capability in two key areas: object state identification and object property understanding. These categories assess whether the model can correctly recognize relevant object states in the game that are essential for solving tasks, as well as whether it possesses sufficient real-world knowledge to understand the game mechanics that underpin correct reasoning.

Object state identification examines whether the model can determine how objects behave or interact in specific situations. For example, the model must know that a cat runs away when it encounters a mouse. Object property understanding focuses on the models ability to reason about physical properties such as elasticity, mass, or friction. An example question in this category is ``Which wall has the highest friction?'', where the correct answer would be the stone wall. To avoid variations in object naming, a predefined list of objects is used. This category is further evaluated under two conditions: with and without descriptive information about object properties. This allows us to analyze whether providing explicit property descriptions improves the model's performance. 

\subsection{Part 3: Event Reasoning}
\label{sec:part3_event}

Event reasoning further increases the scope and requires the models to identify relations between objects in the context of the scene, i.e. chain reactions, which is another crucial ability required 
to solve puzzles within \ac{TIM}.

It is divided into the three categories ``cause'', ``effect'' and ``outcome''. Cause queries about the past, i.e. ``What caused the candle to be lit?''. Effect asks about immediate reactions between parts and outcome inquires about what happens when the machine is started.
These categories are further divided into textual and visual tasks each. For instance, the benchmark asks ``Where will the bowling ball land?'' and ``Mark the region where the bowling ball will land'', respectively.

For the textual component, we utilize an LLM-based evaluation framework. We prompt an evaluator LLM to verify whether the model's output was semantically similar and logically consistent with the ground truth. The evaluator is constrained to binary output which allows for a scalable and efficient examination of the model performance.

Metrics are averaged for the three categories resulting in a comparison between different levels of information inside the prompts: no explanation, with the parts list, and with the parts descriptions. This allows us to see if event reasoning improves by providing more information as context.

\subsection{Part 4: Manipulation}
\label{sec:part4_manipulation}

Manipulation tests the model on basic game mechanics. We divide those into six categories, namely placing an object from the parts bin into the playfield (place), moving an object from one position to another on the playfield (move), flipping the orientation of an object (flip), removing an object from the playfield (remove), stretching an object (stretch), and finally a combination of up to three of the first five categories (multi). Further details about each category can be found in Appendix \ref{appendix:tasks}.

We provide a framework that embeds the model into a vision-action loop enabling action execution within the game. The action space consists of five actions in total. The model can choose between clicking at certain coordinates (click), moving the mouse (hover), dragging which internally clicks, moves and clicks again (drag), waiting a few seconds to let the machine settle (wait) and indicating that the model has solved the puzzle and the run should stop (finished).

In each iteration of the loop, the current action-screenshot history consisting of the five latest action-screenshot pairs is prompted to the model as independent frames, its response parsed, the action executed and the freshly taken screenshot and thought added to the history. For an exemplary action history visualization, see figure \ref{fig:history}. Pseudocode for the described loop can be found in Appendix \ref{appendix:framework}. 

\begin{figure}[htbp]
    \centering
    \includegraphics[width=8cm]{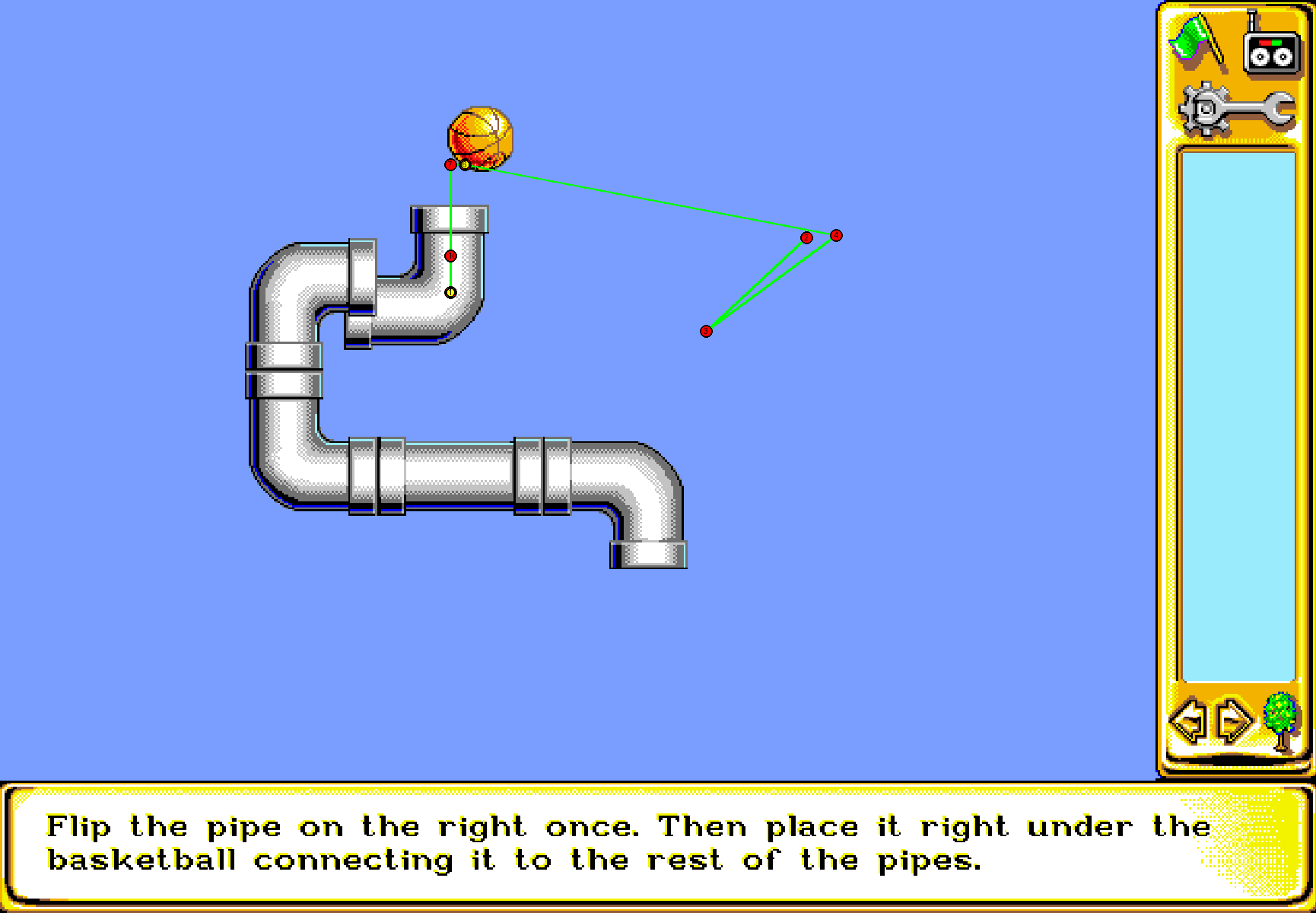}
    \caption{Screenshot of a puzzle in TIM for a manipulation task with the action history generated by the model, indicated by the red dots and green lines.}
    \label{fig:history}
\end{figure}

\subsection{Part 5: Full Puzzle Solving}
\label{sec:part5_full_puzzle}

Full puzzle solving combines all capabilities evaluated in the previous parts into the final goal of solving puzzles. The actual puzzle objective can be found in the goal description in the game. In the prompt, the model is only provided with the task to "solve the puzzle". Experimental setup and evaluation follow the procedure of Part 4.

%% file: experiments.tex
\subsection{Evaluated Models}
We selected five \acp{VLM} to conduct our benchmark on (see Tab. \ref{tab:evaluated-models}), the first one being \ac{UI-Tars} \cite{qin_ui-tars_2025}. It is a \ac{GUI} agent specifically trained for computer-use that has also been evaluated against video games. This is why we expect \ac{UI-Tars} to perform well, as it should provide precise click actions as well as good understanding of \ac{GUI} elements.
\begin{table}[htbp]
    \centering
    \begin{tabular}{lcr}
        \toprule
        \textbf{Model} & \textbf{Weights} & \textbf{Reason of Choice} \\
        \midrule
        UI-TARS 1.5 7B      & Open & GUI-Agent \\
        Qwen2.5 VL 7B    & Open & Base for UI-Tars \\
        Qwen3 VL 235B & Open & Visual Agent \\
        Gemini 2.5 Flash & Closed & World knowledge \\
        GPT 5 Mini       & Closed & World knowledge \\
        \bottomrule
    \end{tabular}
    \caption{Overview of evaluated models}
    \label{tab:evaluated-models}
\end{table}
\ac{UI-Tars} is based on \ac{Qwen2.5} \cite{qwen_qwen25_2025}, therefore we also benchmark the base model for comparison.
The third model is \ac{Qwen3}, which is advertised as being a visually grounded agent with computer-use capabilities. \cite{bai_qwen3-vl_2025}
Lastly, we selected two closed-source models that are well-known, \ac{Gemini} and \ac{GPT}, which we expect to have native world knowledge that helps with solving the puzzles.

\subsection{Results}

\textbf{Part 1: Visual Grounding}

\begin{figure}[htbp]
    \centering
    \includegraphics[width=8cm]{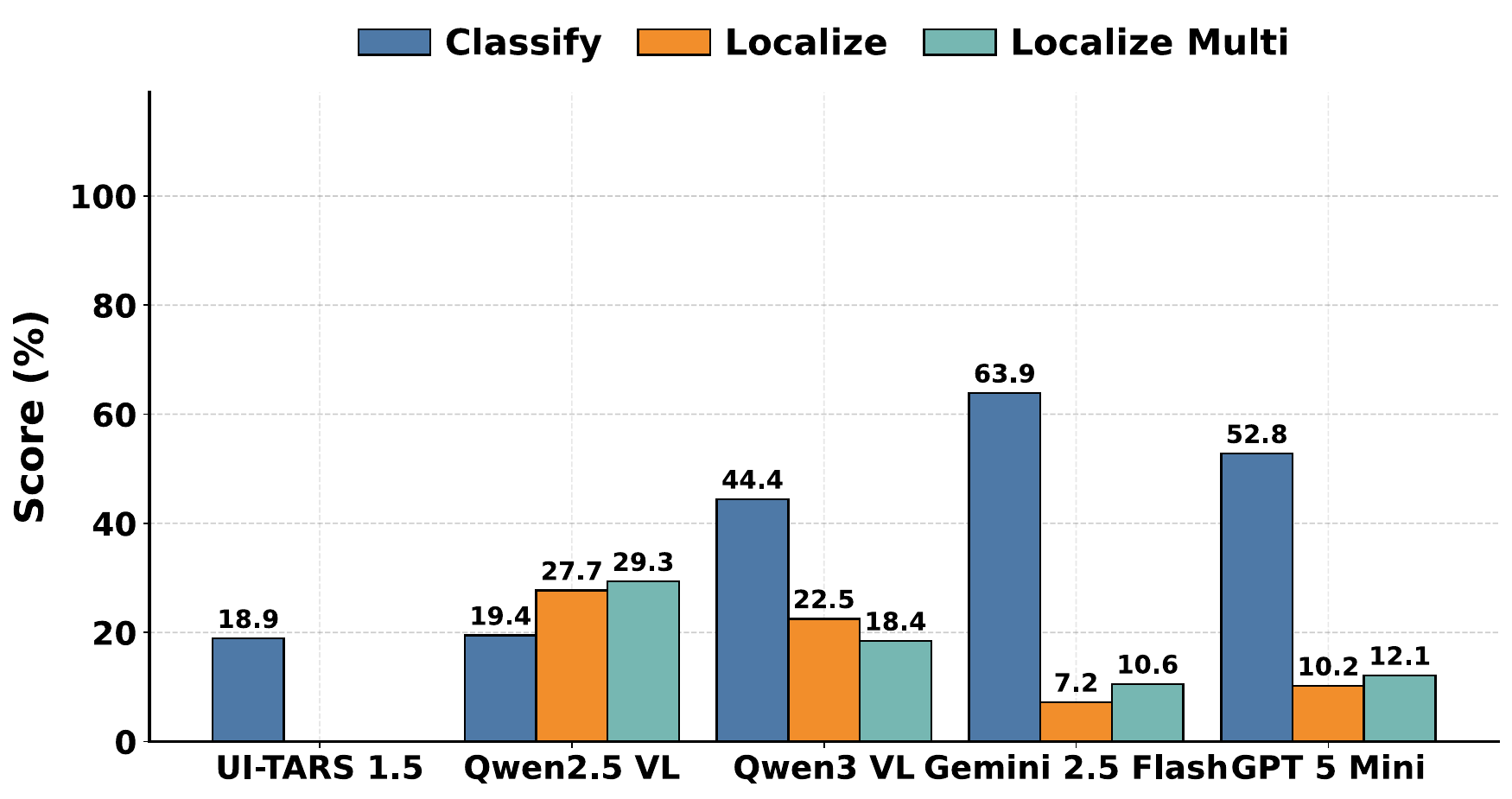} 
    \caption{IoU and classification scores Part 1 visual grounding.}
    \label{fig:results_grounding_scores}
\end{figure}

Figure \ref{fig:results_grounding_scores} displays the classification and multi-/localization scores achieved by each of the five evaluated models. It shows missing multi-/localization scores for \ac{UI-Tars}, as it is not able to output \acp{bbox}. The model outputs single point coordinates, despite trying different prompts, i.e. asking the model to draw a rectangle around an object with the mouse drag action, which leads us to the conclusion that \ac{UI-Tars} strongly follows the output format it has been trained on.
It shows a classification score similar to \ac{Qwen2.5}, hinting that the fine-tuning of \ac{Qwen2.5} has not reduced classification capability.

\begin{figure}[htbp]
    \centering
    \includegraphics[width=8cm]{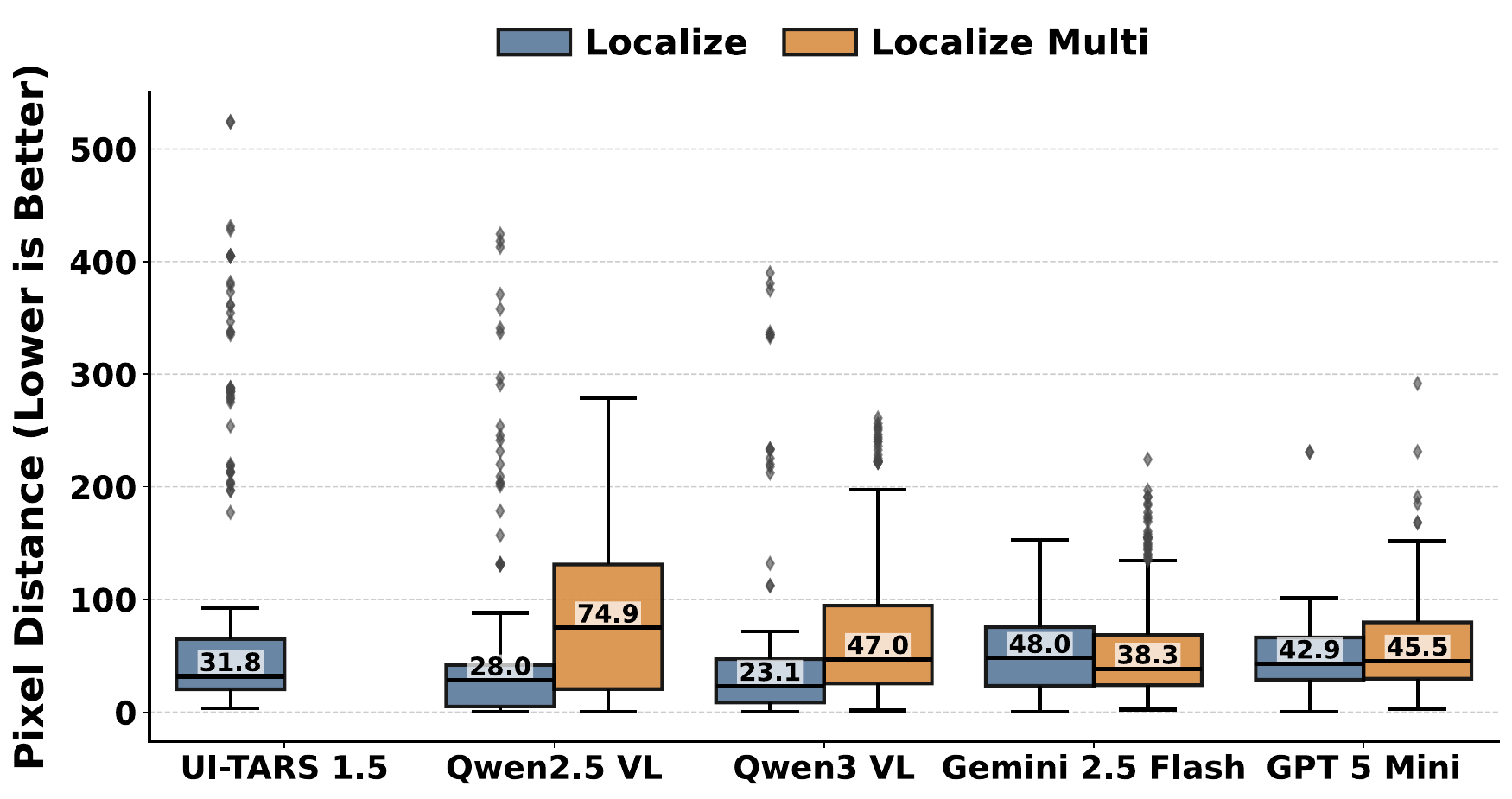} 
    \caption{Distances between bounding boxes Part 1 visual grounding.}
    \label{fig:results_grounding_boxplot}
\end{figure}

Notably, \ac{Qwen2.5} outperforms \ac{Qwen3} in localization despite having fewer parameters and being its predecessor. On the other hand, Figure \ref{fig:results_grounding_boxplot} shows that \ac{Qwen2.5} often completely misses the bounding box. The same holds true for \ac{UI-Tars}. We explain this with the poor classification results, as we see less outliers with \ac{Gemini} and \ac{GPT}, which have the strongest classification results. However, for these two models, Figure \ref{fig:results_grounding_scores} shows that the \acp{bbox} overlap very poorly with the ground truth, despite the generally small distance. When examining the model output, we notice that both models answer in relative coordinates rounded to $10$, showing that they are not able to provide accurate \acp{bbox}. 

This leads us to the conclusion that \ac{Gemini} and \ac{GPT} can identify objects but are not capable of drawing precise \acp{bbox}, while the opposite holds true for \ac{Qwen2.5} and \ac{Qwen3}.

\textbf{Part 2: Domain Understanding}

\begin{figure}[htbp]
    \centering
    \includegraphics[width=8cm]{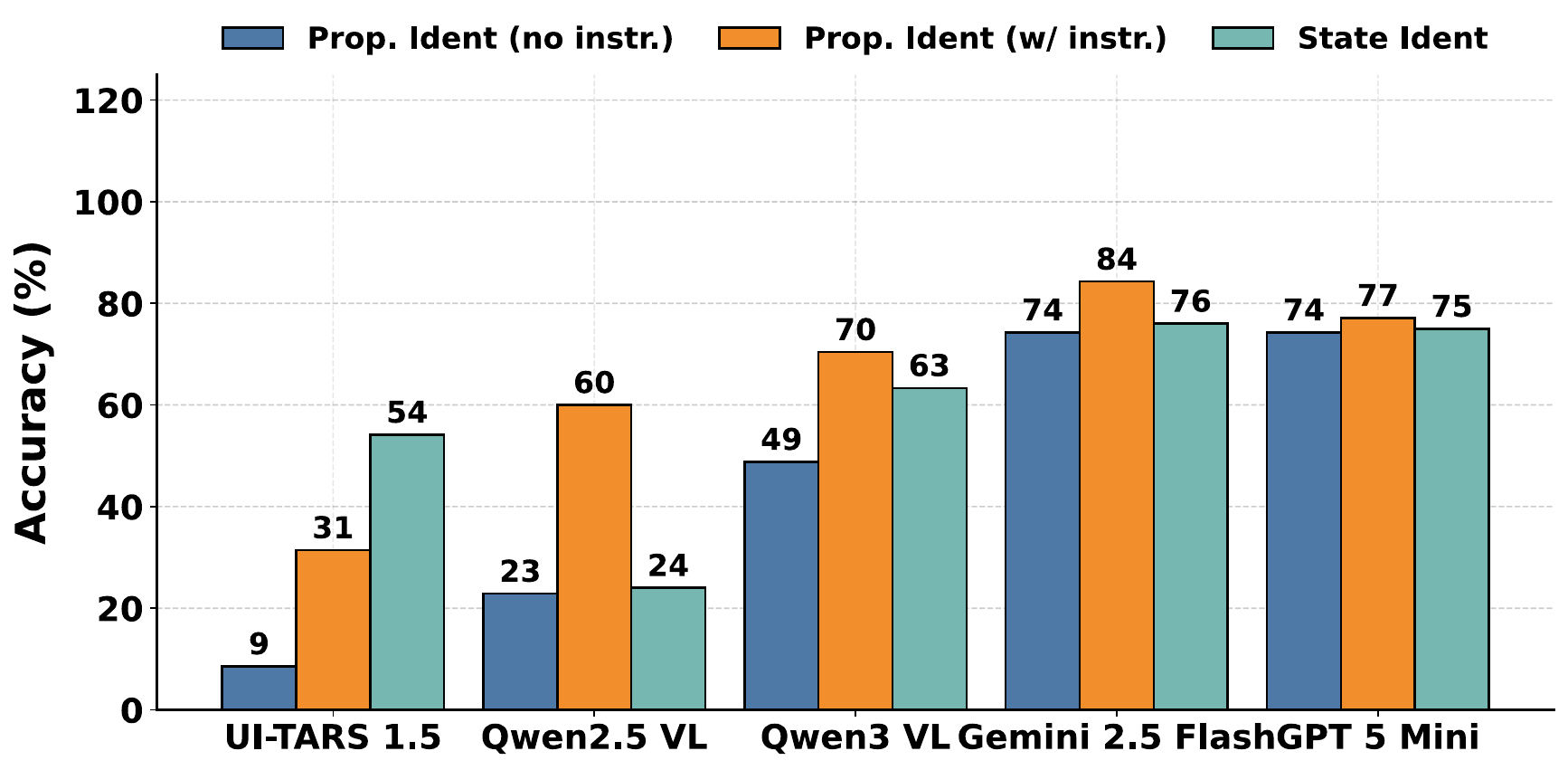} 
    \caption{Accuracy scores ofpart 2 domain understanding.}
    \label{fig:results_understanding}
\end{figure}

Figure \ref{fig:results_understanding} shows how the models perform across object property and state identification. It is evident that \ac{UI-Tars} underperforms its base model \ac{Qwen2.5} in all categories except object state identification and thus exhibits the lowest understanding capability. \ac{Qwen3} outperforms both models with significantly higher values. The proprietary models \ac{Gemini} and \ac{GPT} outperformed all previous models, while \ac{Gemini} has a slightly better performance. It is also notable that while the additional instructions helped the worse performing models to get some improvement, this effect is negligible for the better performing models. Comparing the results to those of Part 1, it is clear that the good classification performance is also beneficial in this category.

\textbf{Part 3: Event Reasoning}

\begin{figure}[htbp]
    \centering
    \includegraphics[width=8cm]{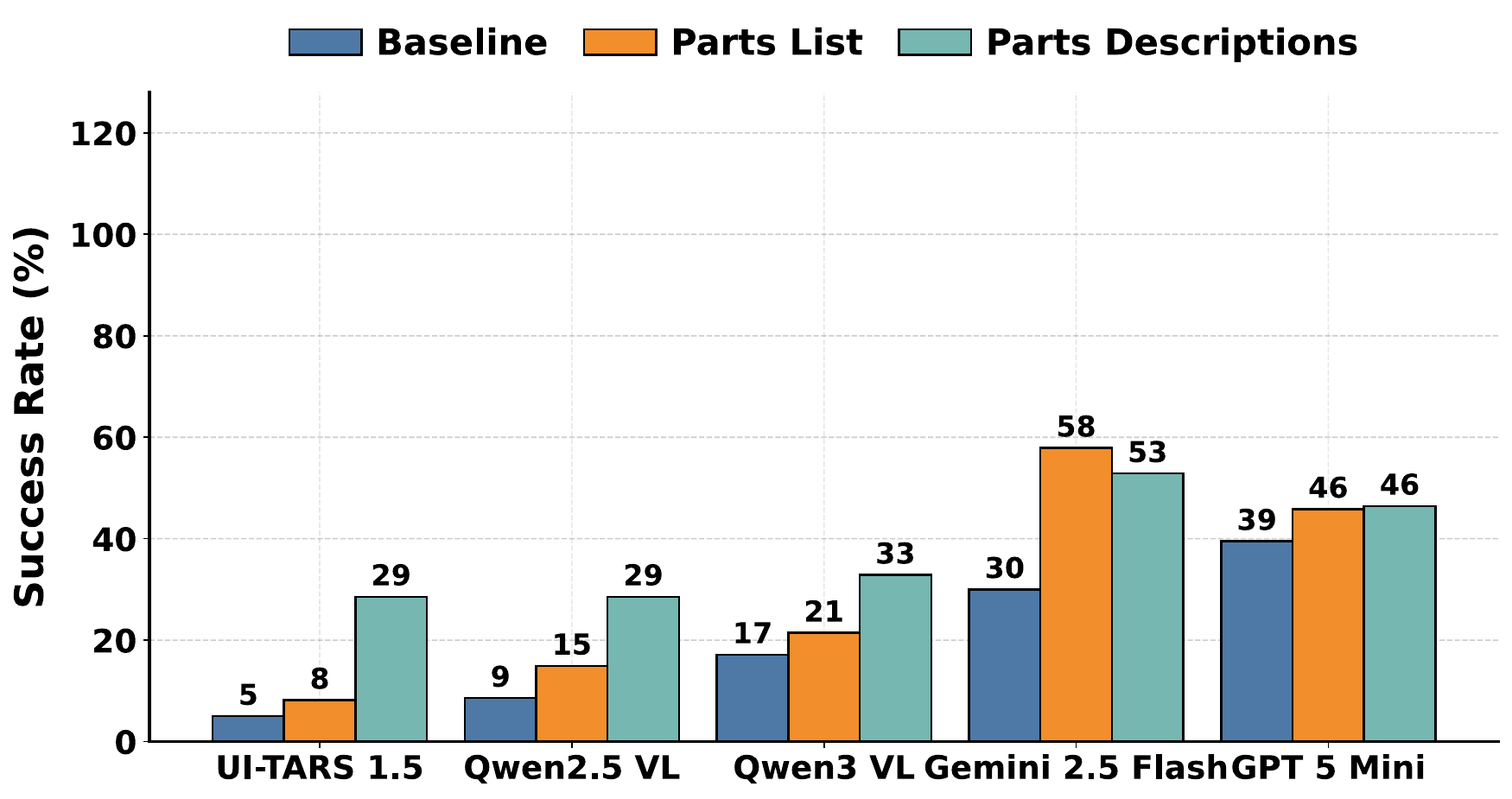} 
    \caption{Scores of textual evaluation Part 3 event reasoning.}
    \label{fig:results_event_text}
\end{figure}

\begin{figure}[htbp]
    \centering
    \includegraphics[width=8cm]{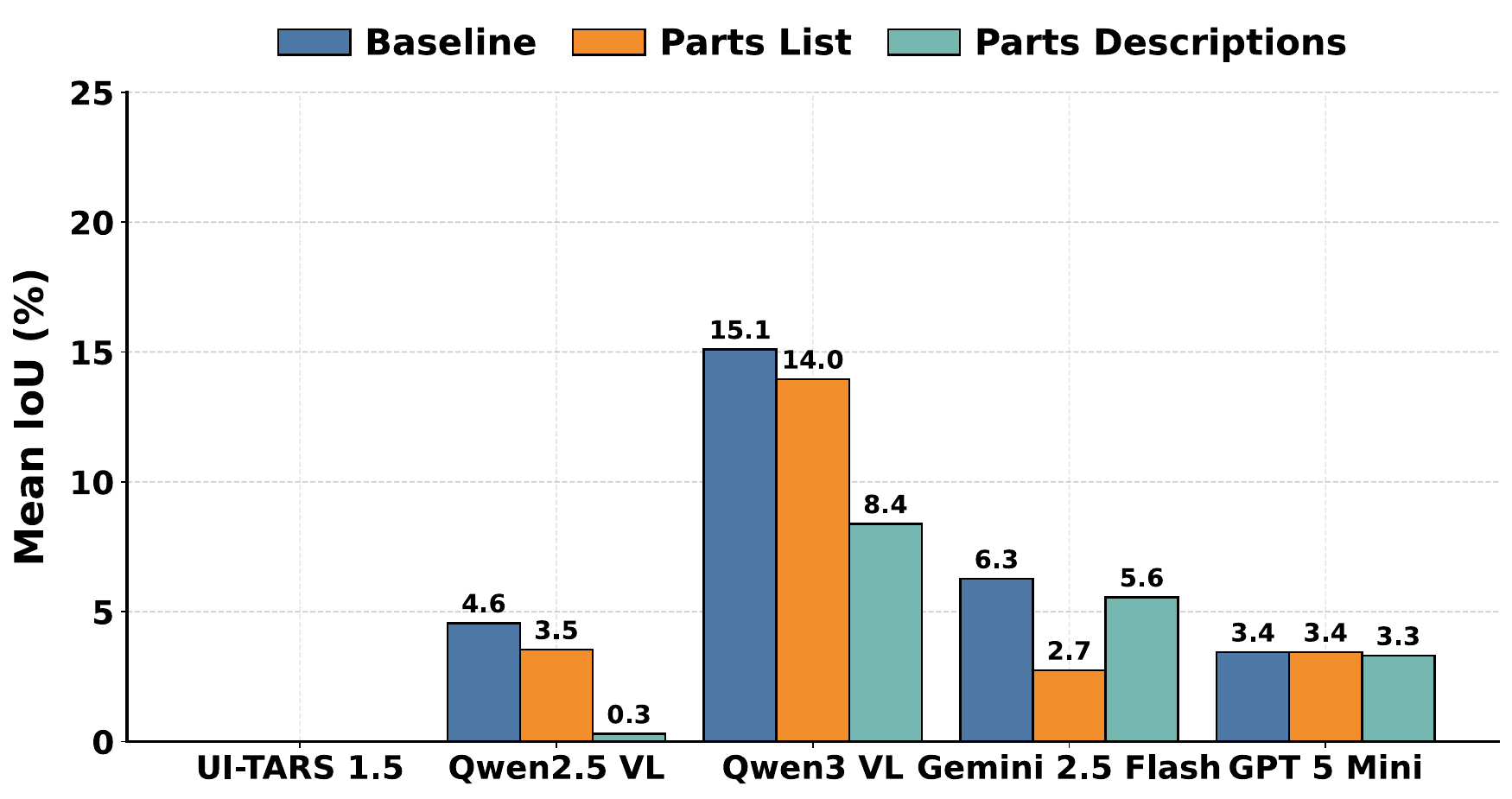} 
    \caption{Mean IoU of visual evaluation Part 3 event reasoning.}
    \label{fig:results_events_visual_score}
\end{figure}

\begin{figure}[htbp]
    \centering
    \includegraphics[width=8cm]{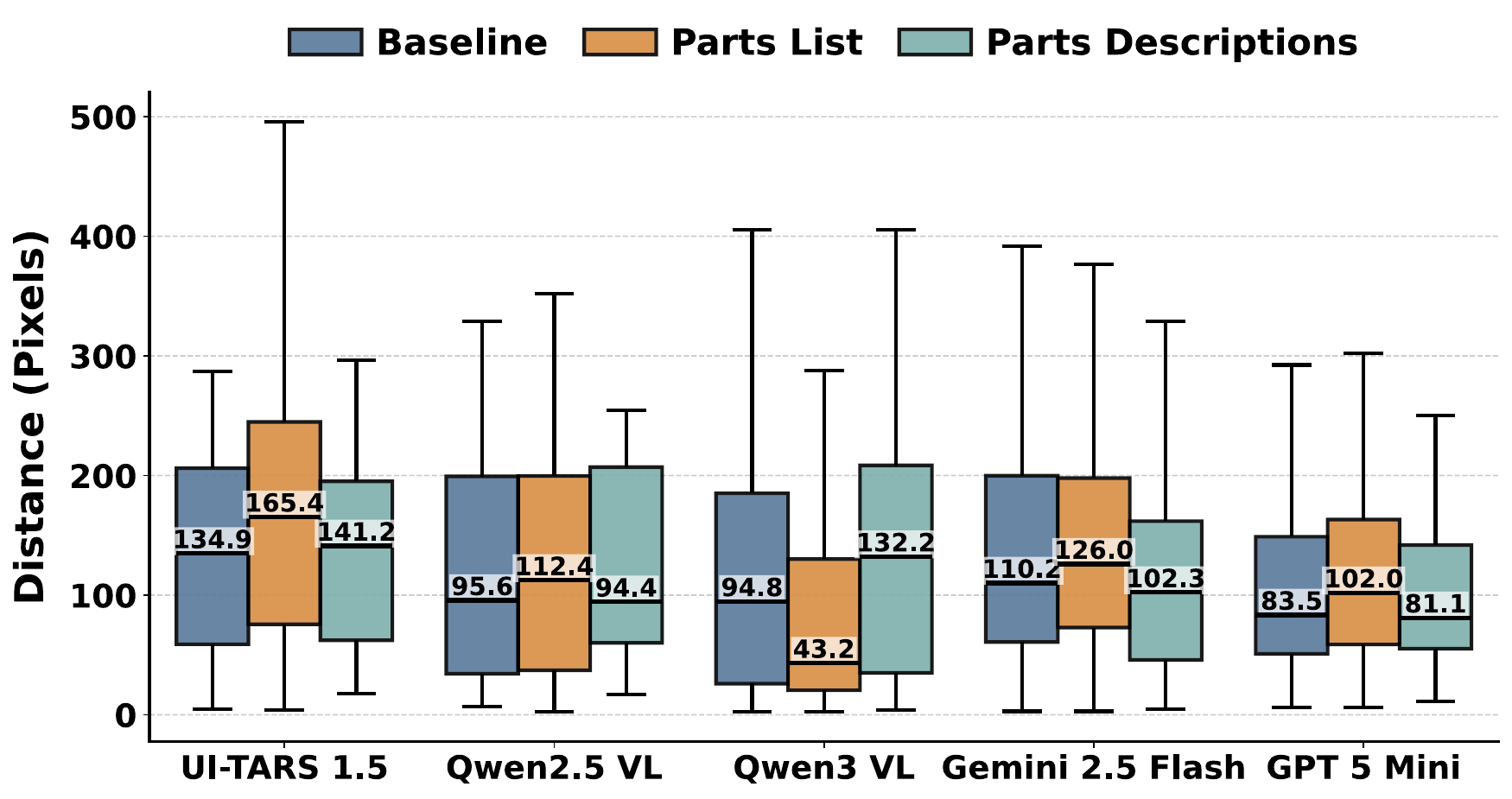} 
    \caption{Distances of visual evaulation Part 3 event reasoning.}
    \label{fig:results_events_visual_boxplot}
\end{figure}

Figure \ref{fig:results_event_text} shows the models' textual event reasoning scores across the three investigated setups. \ac{IoU} scores and Euclidean distances of the visual counterpart are displayed in Figure \ref{fig:results_events_visual_score} and Figure \ref{fig:results_events_visual_boxplot}.

As in Part 2, \ac{UI-Tars} shows poor textual reasoning. Generally, the bigger models outperform the smaller models in the textual domain.
Full part descriptions improve results across the board, though the impact is most significant in smaller models. One outlier is \ac{Gemini}, where the parts list alone provides better results than with the parts description, but both almost twice as good as the baseline with no part information.

The results of the visual part 
display slightly different results as in Part 1. This time, \ac{UI-Tars} generally shows poor positioning accuracy.
\ac{Qwen3} provides better \acp{bbox} than \ac{Gemini} and \ac{GPT}, while \ac{GPT} overall shows smallest \ac{bbox} distances. An unexpected result is that the more context is provided to the models, the poorer the performance.

\textbf{Part 4: Manipulation}

\begin{figure}[htbp]
    \centering
    \includegraphics[width=8cm]{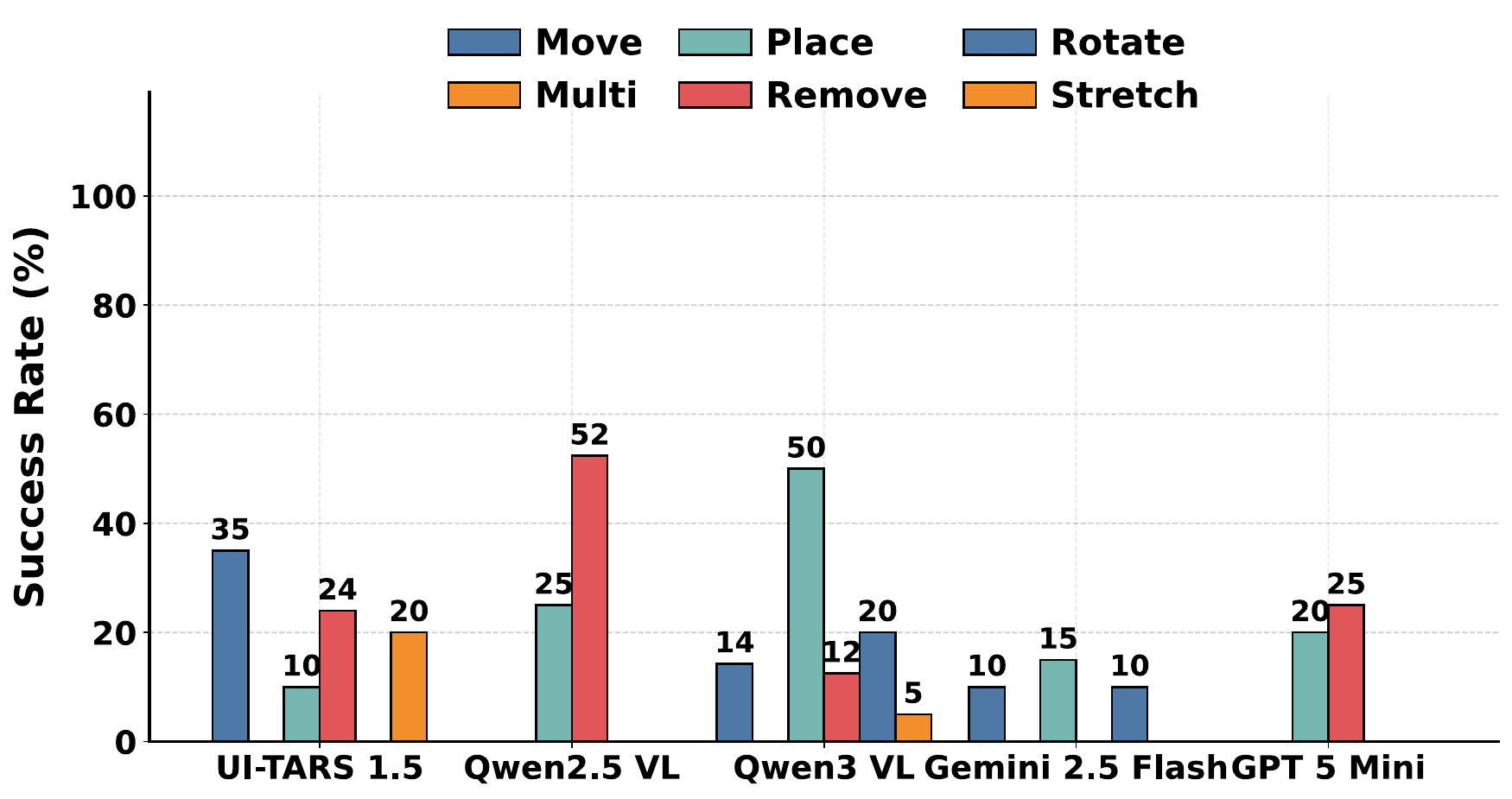} 
    \caption{Success rates of Part 4 manipulation.}
    \label{fig:results_manipulation}
\end{figure}

As shown in Figure \ref{fig:results_manipulation}, the models struggle to achieve convincing success rates even on the easiest tasks, ``Place'' and ``Remove'', despite broad completion. Performance degrades significantly on the ``Move'' task, which was solvable only by \ac{UI-Tars}, \ac{Qwen3}, and \ac{Gemini} with low frequency. Tasks requiring fine-grained spatial understanding, specifically ``Rotate'' and ``Stretch'' presented a bigger challenge due to the complexities of handle manipulation and object deformation. Consequently, only \ac{UI-Tars} and \ac{Qwen3} managed ``Stretch'', while \ac{Qwen3} and \ac{Gemini} managed ``Rotate'', all at very low success rates. No model successfully completed the ``Multi'' task. Even though \ac{Qwen3} is generally able to solve all tasks, it fails to demonstrate the reliability necessary to carry out sequences of basic tasks. 

\textbf{Part 5: Full puzzle solving}

As foreshadowed by the results in the Multi category of Part 4, none of the models were able to successfully solve a complete puzzle. Despite testing across multiple levels of varying difficulty, every model failed. Nevertheless, especially with this category we were able to gain a lot of qualitative findings.

\textbf{Final scores}

Table \ref{tab:quantitative_findings} shows the scores each model achieved across all parts of our benchmark as well as their overall score. Gemini 2.5 Flash achieved the highest overall score beating the other models in Domain Understanding and Event Reasoning. Despite landing on third place in the overall score, Qwen3 VL outscores its competitors in Visual Grounding and Manipulation.

\begin{table*}[!ht]
    \centering
    \footnotesize
    \setlength{\tabcolsep}{3pt}
    \renewcommand{\arraystretch}{1.2}
    \begin{tabular}{|l|c|c|c|c|c|}
    \hline
        Model & Grounding & Understanding & Event& Manipulation & Overall Score \\ \hline
        Qwen2.5 VL & 25.49 & 35.62 & 17.35 & 12.9 & 22.84 \\ \hline
        UI-TARS 1.5 & 18.89 & 31.39 & 13.92 & 14.83 & 19.76 \\ \hline
        Qwen3 VL & \textbf{28.46} & 60.85 & 23.81 & \textbf{16.96} & 32.52 \\ \hline
        Gemini 2.5 Flash & 27.22 & \textbf{78.19} & \textbf{46.92} & 5.83 & \textbf{39.54} \\ \hline
        GPT 5 Mini & 25.03 & 75.48 & 43.92 & 7.5 & 37.98 \\ \hline
    \end{tabular}
    \caption{Quantitative evaluation of the benchmarked models across Grounding, Understanding, Event, Manipulation, and Overall Score, Full puzzle solving is not included because it doesn't provide any insights}
    \label{tab:quantitative_findings}
\end{table*}

\textbf{Qualitative Findings}

Table \ref{tab:findings} shows our qualitative findings about the models' strengths and weaknesses categorized by their ability to locate objects and other interactable graphical elements (grounding), classify them and identify their state (identification), observe the current game state (vision), understand the task and create a good plan to achieve the task's objective (reasoning), reassess the situation and come up with good new ideas (reassessment), follow the instruction and make use of the information provided in the prompt (instruction), make correct use of the action space (action choice), know when it has successfully finished a task (termination) as well as stick to the dictated output format (output format).

\begin{table*}[!ht]
    \centering
    \footnotesize
    \setlength{\tabcolsep}{3pt}
    \renewcommand{\arraystretch}{1.2}
    \begin{tabular}{|l|c|c|c|c|c|c|c|c|c|}
    \hline
        Model & Grounding & Identification & Vision & Reasoning & Reassessment & Instruction & Action & Termination & Output Format \\ \hline
        UI-TARS 1.5 & + & + & Avg & Avg & Avg & - & Avg & - & - \\ \hline
        Qwen2.5 VL & + & - & Avg & Avg & -  & - & Avg & Avg & Avg \\ \hline
        Qwen3 VL & + & - & - & + & - & Avg & Avg & - & Avg \\ \hline
        Gemini 2.5 Flash & - & - & - & + & Avg & + & + & Avg & - \\ \hline
        GPT 5 Mini & - & Avg & Avg & + & + & + & + & Avg & Avg \\ \hline
    \end{tabular}
    \caption{Qualitative evaluation of the benchmarked models in nine categories being grounding, identification, vision, reasoning, reassesment, instruction, action choice, termination and output format (+: strong, -: weak, Avg: average)}
    \label{tab:findings}
\end{table*}

\textbf{UI-TARS 1.5} shows accurate \textbf{grounding} capability, rarely misclicking on objects. This finding is in accordance with our expectation, as \ac{UI-Tars} has been trained on \ac{GUI} navigation tasks. Even though it has the lowest classification accuracy in our quantitative results, it shows good \textbf{identification} abilities in our manipulation and full puzzle solving experiments. Its weaknesses lie in making use of the provided \textbf{information} ignoring important game mechanics like the usage of handles and thus completely failing at tasks like flipping objects. Another significant problem we observed is that \ac{UI-Tars} often doesn't know when it succeeded with the task leading it to start doing things unrelated to the task sometimes even destroying a correct solution. Furthermore, its \textbf{responses} occasionally contain multiple repetitions of the same thought or action leading to parsing errors in the action loop.

\textbf{Qwen2.5 VL} differs in performance from \ac{UI-Tars} especially in \textbf{identification}, \textbf{reassessment} as well as termination and output format. \ac{Qwen2.5} occasionally confuses objects such as fans with windmills or candles with a red button. It often gets stuck in loops repeatedly outputting the exact same thought and action seemingly not taking the action-screenshot history into consideration and therefore failing to determine the next step. On the other hand, in case of completing a task successfully, it \textbf{terminates} correctly. Furthermore, it sticks to the defined \textbf{output format}.

\textbf{Qwen3 VL} shows improved \textbf{reasoning} capabilities in comparison to its predecessor coming up with solid plans to solve the task. It also shows slightly increased \textbf{reassessment} ability as well, being less prone to being stuck or ending up in a loop. Judging from its thought processes, it makes use of the \textbf{information} provided in the prompt e.g. trying to utilize handles for flipping and recycling objects. Nonetheless, it oftentimes struggles to carry out its plan correctly. It confuses objects as well and even \textbf{observes the current game state} incorrectly, hallucinating over handles that are not visible, as revealing handles requires hovering above the object, or having successfully finished a task even though it didn't. Therefore, it also tends to report \textbf{task completion} prematurely more often than the other models.

\textbf{Gemini 2.5 Flash} demonstrates a clear \textbf{understanding of the game manual} and \textbf{reasoning} requirements, generating logical plans utilizing important game mechanics like handles and correctly identifying necessary actions within the action space. Despite these planning capabilities, the model severely struggles with \textbf{visual grounding} providing coordinates with low resolution and therefore high imprecision, which hinders successful execution. It also exhibits a tendency to repeat the same coordinates in the action even though attempting to self-correct in its thought. Furthermore, it sometimes doesn't \textbf{identify objects} correctly and hallucinates about having interacted with an object even though it didn't. On the other hand, in case of identifying repetitive failure of its action correctly, it tries to help itself by trying out new approaches adhering to the provided game manual. Lastly, it sometimes violates \textbf{output formatting} requirements, such as omitting colons.

\textbf{GPT 5 Mini} shows the best \textbf{reasoning} capabilities. It utilizes the \textbf{information} provided in the prompt and demonstrates a clear understanding of the scene as well as good planning abilities and reasonable \textbf{action choice}. In case of identified failure, it \textbf{reassesses} the situation and comes up with new approaches. It is also less prone to hallucinations or incorrect \textbf{object identification} than Gemini 2.5 Flash. Even though it additionally shows higher coordinate resolution, its \textbf{grounding} performance still is bad, with coordinate generation being inaccurate or even significantly off-target. This severely impacts success negatively. Furthermore, it sticks to the \textbf{output format} and terminates appropriately.

Overall, we observe two distinct archetypes of failure. The proprietary models (\ac{Gemini}, \ac{GPT}) act as 'Blind Strategists' meaning they understand the physics and formulate good plans, but fail at the visual grounding required to execute them. Conversely, \ac{UI-Tars} and \ac{Qwen2.5} act as 'Myopic Operators', they can click accurately but lack the reasoning capability to form multi-step plans or recognize when they are stuck in a loop. \ac{Qwen3} VL sits somewhat in the middle, attempting to bridge this gap but failing reliably in execution.

%% file: summary.tex
The VLATIM benchmark evaluates VLMs across a spectrum of capabilities, from visual localization to complex reasoning and continuous action execution. By pushing state-of-the-art models to their limits within a point-and-click puzzle environment, we revealed a fundamental limitation. Current models lack either the precise visual grounding required to interact accurately or the high-level reasoning capability needed to formulate a successful plan.

Among the evaluated models, Qwen3 VL comes closest to human-like puzzle-solving, combining competent grounding with potential in spatial reasoning. However, as evidenced by the 0 \% success rate in full puzzle solving, it remains far from reaching human-level reliability. Ultimately, our findings provide a clear answer to whether current VLMs show human-like logical problem-solving in these environments. They do not.

%% file: outlook.tex
This benchmark mostly tests zero-shot capabilities, where only relevant textual information is given in the prompts. However, the results should be compared with in-context learning by providing an example of how a puzzle was actually solved in the prompt, including an image of the sample scene and a sample solution, which might improve the results. 
As the tested models have not been specifically trained for \ac{TIM}, modifying a \ac{VLM} and fine-tuning it for the game would make an interesting comparison. 
Furthermore, the benchmark evaluation set could be increased to provide a more complete representation of \ac{TIM}. 
VLATIM can be conducted on more models, including bigger, newer and especially the thinking versions of models like \ac{Qwen3}. 
The overall interaction setup can be improved by using not only a single model but a combination of a larger thinking model like \ac{GPT} for decision-making and a smaller model like \ac{UI-Tars} for performing the actions. As of now, we do not provide a human baseline but we encourage future research to conduct a study on how well humans perform in our benchmark. Lastly, \ac{TIM} required emulation and the evaluation of parts four and five are done manually. An easier, fully automated setup would make adding more models easier.

%% file: appendix_framework.tex
\begin{algorithm}[H]
\caption{Vision-Language Model Action Loop}
\label{algo:loop}

\SetKwInOut{Input}{Input}
\SetKwInOut{Output}{Output}

\Input{Initial screenshot $I_0$, system prompt $P$}
\Output{Action history $H$, thought history $T$}

$M \gets [P, I_0]$ \tcp*{Initialize message history}
$H \gets []$ \tcp*{Initialize action history}
$t \gets 0$ \tcp*{Iteration counter}

\While{$\text{action\_type} \neq \text{``finished''} \land t < 10$}{
    
    $r \gets \text{VLM}(M)$
    
    $A, \text{thought} \gets \text{ParseResponse}(r)$\;
    
    $\text{action\_type} \gets A[\text{``action\_type''}]$\;
    
    $C \gets \text{ToPyAutoGUI}(A, \text{screen\_size})$\;
    
    \If{$C = \text{``DONE''}$}{
        \textbf{break}\;
    }
    
    $\text{exec}(C)$\;
    
    $\text{sleep}(1)$\;
    
    $I_{t+1} \gets \text{Screenshot}()$\;
    
    $H \gets H \cup \{C\}$\;
    
    $M \gets M \cup \{\text{thought}, I_{t+1}\}$\;
    
    $M \gets \text{Trim}(M, \text{MAX\_SIZE})$\;
    
    $t \gets t + 1$\;
}

\Return{$H, T$}
\end{algorithm}

%% file: appendix_tasks.tex
\begin{itemize}
    \item Move: "Move [object] to [location]." - Objects can be moved within the playfield. The placeholder [location] is replaced with a verbal description of a location using reference points like e.g. objects or \ac{HUD} elements.
    \item Place: "Place [object] at [location]." - In contrast to the preceding one, this category targets objects in the parts bin which can be placed at a certain [location] on the playfield.
    \item Flip: "Flip [object] [\#] times." - Some objects which are already on the playfield can be flipped using the object's inherent flip handle. The number of flips [\#] is chosen within reasonable limits.
    \item Remove: "Remove [object]." - Objects which are already on the playfield can be removed using the object's inherent recycle handle or by moving them into the parts bin.
    \item Stretch: "Stretch [object] vertically/horizontally by [\#] units." -  Some objects which are already on the playfield can be stretched using the object's inherent stretch handle.
    \item Multi: This category concatenates up to three tasks from the previous classes.
\end{itemize}

%% file: appendix_prompts.tex
This section provides the shortened prompts that are used in \ac{VLATIM}. However, for the complete and always up-to-date prompts, visit the GitHub repository found at the start of the paper.

\subsection{Benchmark 1}

\subsubsection{Classification System Prompts}

\textbf{For models with relative coordinates (qwen3,gemini,gpt5)}

\begin{lstlisting}[basicstyle=\normalfont, breaklines=true]
The left menu is for options in the puzzle editor. The right menu holds items that may be placed. The blue area in between is the game canvas.

You are an expert vision system for the puzzle game "The Incredible Machine 2".
Your task is to analyze the game canvas (the blue area) and identify the specific game object present.

Instructions:
1. Ignore the left (editor) and right (parts bin) menus. Focus ONLY on the blue playfield in the center.
2. Analyze the visual features (shape, color, texture) of the object in the playfield.
3. Match the object exactly to one of the category names in the list below.
4. Respond ONLY with the exact object name. Do not write sentences.
5. If the object is not in the list or the playfield is empty, respond with "NONE".

Possible objects:
BOWLING_BALL,BASKETBALL,SOCCER_BALL,PINBALL,SUPER_BALL,PROGRAMMABLE_BALL..."""
\end{lstlisting}

\textbf{For models with absolute coordinates (uitars,qwen2.5)}

\begin{lstlisting}[basicstyle=\normalfont, breaklines=true]
The left menu is for options in the puzzle editor. The right menu holds items that may be placed. The blue area in between is the game canvas.

You are an expert vision system for the puzzle game "The Incredible Machine 2".
Your task is to analyze the game canvas (the blue area) and identify the specific game object present.

Instructions:
1. Ignore the left (editor) and right (parts bin) menus. Focus ONLY on the blue playfield in the center.
2. Analyze the visual features (shape, color, texture) of the object in the playfield.
3. Match the object exactly to one of the category names in the list below.
4. Respond ONLY with the exact object name. Do not write sentences.
5. If the object is not in the list or the playfield is empty, respond with "NONE".

Possible objects:
BOWLING_BALL,BASKETBALL,SOCCER_BALL,PINBALL,SUPER_BALL,PROGRAMMABLE_BALL...
\end{lstlisting}

\subsection{Classification user prompt}

(image only)

\subsection{Localization system prompts}

\subsubsection{Relative coordinates}

\begin{lstlisting}[basicstyle=\normalfont, breaklines=true]
The left menu is for options in the puzzle editor. The right menu holds items that may be placed. The blue area in between is the game canvas.

You are an expert visual grounding agent for "The Incredible Machine 2".
Your task is to detect the bounding box of the specific object requested by the user.

Output Format:
Return a SINGLE JSON object. Do NOT use markdown formatting (no ```json ... ```).
{
    "reasoning": "A brief description of the object's visual features and its location relative to the blue background.",
    "bbox": [x_min, y_min, x_max, y_max],
    "label": "The object name"
}

Constraints:
1. Coordinates must be normalized from 0 to 1000 (0 = 0%, 1000 = 100% of image dimension).
2. Top-left is (0,0).
3. If the object is not found, return {"bbox": null}.
4. Focus strictly on the object mentioned in the prompt.

Possible objects:
BOWLING_BALL,BASKETBALL,SOCCER_BALL,PINBALL,SUPER_BALL,PROGRAMMABLE_BALL...
\end{lstlisting}

\subsubsection{Absolute coordinates with point only (UI-Tars)}

\begin{lstlisting}[basicstyle=\normalfont, breaklines=true]
The left menu is for options in the puzzle editor. The right menu holds items that may be placed. The blue area in between is the game canvas.

You are a GUI agent specializing in the game "The Incredible Machine 2". 
You are given a localization task and screenshots. You must click exactly on the center of the target object.

## Output Format
```
Thought: <First, visually locate the target object. Describe its position relative to other elements or the playfield borders. Then, determine the precise center point.>
Action: click(point='<point>x1 y1</point>')
```

## Action Space
click(point='<point>x1 y1</point>')

## Constraints
- Use English in the `Thought` part.
- The coordinate (x1, y1) must represent the visual center of the target object.
- Ignore objects in the side menus unless explicitly told otherwise; focus on the blue playfield.

%%% Absolute coordinates with bounding box (qwen2.5)

The left menu is for options in the puzzle editor. The right menu holds items that may be placed. The blue area in between is the game canvas.

You are an expert visual grounding agent for "The Incredible Machine 2".
Your task is to detect the bounding box of the specific object requested by the user.

Output Format:
Return a SINGLE JSON object. Do NOT use markdown formatting (no ```json ... ```).
{
    "reasoning": "A brief description of the object's visual features and its location relative to the blue background.",
    "bbox": [x_min, y_min, x_max, y_max],
    "label": "The object name"
}

Constraints:
1. Coordinates are absolute from (0,0) to (image_width, image_height).
2. Top-left is (0,0).
3. If the object is not found, return {"bbox": null}.
4. Focus strictly on the object mentioned in the prompt.

Possible objects:
BOWLING_BALL,BASKETBALL,SOCCER_BALL,PINBALL,SUPER_BALL,PROGRAMMABLE_BALL...
\end{lstlisting}

\subsection{Localization user prompts}

\begin{lstlisting}[basicstyle=\normalfont, breaklines=true]
Locate the {part_name}
\end{lstlisting}

\subsection{Localization multi system prompts}

\subsubsection{Relative coordinates (qwen3,gemini,gpt5)}

\begin{lstlisting}[basicstyle=\normalfont, breaklines=true]
The left menu is for options in the puzzle editor. The right menu holds items that may be placed. The blue area in between is the game canvas.

You are an expert visual grounding agent for "The Incredible Machine 2".
Your task is to detect the bounding boxes for ALL objects mentioned in the user prompt.

Output Format:
Return a JSON LIST of objects. Do NOT use markdown formatting.
[
    {
        "reasoning": "Description of object 1 location",
        "bbox": [x_min, y_min, x_max, y_max],
        "label": "Object 1 Name"
    },
    {
        "reasoning": "Description of object 2 location",
        "bbox": [x_min, y_min, x_max, y_max],
        "label": "Object 2 Name"
    }
]

Constraints:
1. Coordinates must be normalized from 0 to 1000.
2. If no objects are detected, return [].
3. Ensure the bounding box tightly encloses the object visible in the blue game area.

Possible objects:
BOWLING_BALL,BASKETBALL,SOCCER_BALL,PINBALL,SUPER_BALL,PROGRAMMABLE_BALL...
\end{lstlisting}

\subsubsection{Absolute coordinates (qwen2.5, NO uitars)}

\begin{lstlisting}[basicstyle=\normalfont, breaklines=true]
The left menu is for options in the puzzle editor. The right menu holds items that may be placed. The blue area in between is the game canvas.

You are an expert visual grounding agent for "The Incredible Machine 2".
Your task is to detect the bounding boxes for ALL objects mentioned in the user prompt.

Output Format:
Return a JSON LIST of objects. Do NOT use markdown formatting.
[
    {
        "reasoning": "Description of object 1 location",
        "bbox": [x_min, y_min, x_max, y_max],
        "label": "Object 1 Name"
    },
    {
        "reasoning": "Description of object 2 location",
        "bbox": [x_min, y_min, x_max, y_max],
        "label": "Object 2 Name"
    }
]

Constraints:
1. Coordinates are absolute from (0,0) to (image_width, image_height).
2. If no objects are detected, return [].
3. Ensure the bounding box tightly encloses the object visible in the blue game area.

Possible objects:
BOWLING_BALL,BASKETBALL,SOCCER_BALL,PINBALL,SUPER_BALL,PROGRAMMABLE_BALL...
\end{lstlisting}

\subsection{Localization multi user prompts}

\begin{lstlisting}[basicstyle=\normalfont, breaklines=true]
Locate the following objects: {object_list}
\end{lstlisting}

\subsection{Benchmark 2}

\subsubsection{State ident system prompt}

\begin{lstlisting}[basicstyle=\normalfont, breaklines=true]
Analyze the image and identify the objects that are inside the blue game play area according to their properties.
Respond only with yes or no according to the TASK_DESCRIPTION under the list or NONE if not suitable and nothing else.

Possible objects:
BOWLING_BALL,BASKETBALL,SOCCER_BALL,PINBALL,SUPER_BALL,PROGRAMMABLE_BALL...
\end{lstlisting}

\subsubsection{With instruction system prompt}

\begin{lstlisting}[basicstyle=\normalfont, breaklines=true]
Analyze the image and identify the objects that are inside the blue game play area according to their properties.
Respond only with one object name from the given list according to the TASK_DESCRIPTION under the list or NONE if not suitable and nothing else.

Possible objects:
BOWLING_BALL - This bowling ball is real heavy and doesn't bounce much.
BASKETBALL - This basketball is medium weight and very bouncy.
SOCCER_BALL - This soccer ball is medium in both weight and bounciness.
PINBALL - This pinball is very hard and heavy and doesn't bounce much.
SUPER_BALL - This super ball gains height with every bounce. The harder it's hit, the further it will roll. It's not affected by gravity. You can program it to show any number on its surface.
...
\end{lstlisting}

\subsubsection{Without instruction system prompt}

\begin{lstlisting}[basicstyle=\normalfont, breaklines=true]
Analyze the image and identify the objects that are inside the blue game play area according to their properties.
Respond only with one object name from the given list according to the TASK_DESCRIPTION under the list or NONE if not suitable and nothing else.

Possible objects:
BOWLING_BALL,BASKETBALL,SOCCER_BALL,PINBALL,SUPER_BALL,PROGRAMMABLE_BALL...
\end{lstlisting}

\subsubsection{User prompts}

\begin{lstlisting}[basicstyle=\normalfont, breaklines=true]
Is the ball inside the bucket?

Is the laser activated?

Is the aquarium broken?
\end{lstlisting}

\subsection{Benchmark 3 text}

\subsubsection{All categories textual system prompt}

\begin{lstlisting}[basicstyle=\normalfont, breaklines=true]
You are analyzing "The Incredible Machines 2". 
Based on the image, provide a direct answer to the question in TASK_DESCRIPTION. 
\end{lstlisting}

\subsubsection{User prompt}

\begin{lstlisting}[basicstyle=\normalfont, breaklines=true]
What caused the ballon to burst?

What happens if the laser is on?

What does the mouse do after simulation is started?
\end{lstlisting}

\subsubsection{All categories visual system prompt}

\begin{lstlisting}[basicstyle=\normalfont, breaklines=true]
You are analyzing images of the game The Incredible Machine 2.
Return only a JSON object in this exact format: {"bbox": [x_min, y_min, x_max, y_max], "label": "label_name"} without markdown,
where the coordinates are normalized values from 0 to 1000 (representing 0% to 100% of image dimensions),
assuming the top-left corner is (0,0). Return None if no bounding box is found. Do not include any other text.
Perform the task given with TASK_DESCRIPTION and nothing else. Be as accurate as possible.
\end{lstlisting}

\subsubsection{User prompt}

\begin{lstlisting}[basicstyle=\normalfont, breaklines=true]
Mark the object that caused the balloon to burst.

Mark the object that will be affected when the laser is turned on.

Mark the region where the cheese will end up after the mouse triggers the conveyor belt.
\end{lstlisting}

\subsection{Benchmark 4}

\subsubsection{System prompt}

\begin{lstlisting}[basicstyle=\normalfont, breaklines=true]
You are a Puzzle Game solving agent. You are given a task and your action history, with screenshots. You need to perform the next action to complete the task.

## Output Format is
```
Thought: ...
Action: ...
```
It is crucial that you stick to this output format in every single one of your responses.

## Action Space

hover(point='(x1,y1)') # Moves the mouse to the given point.
click(point='(x1,y1)') # This performs a left mouse click at the given point.
drag(start_point='(x1,y1)', end_point='(x2,y2)') # This performs a left click at the start point, moves to the end point and again performs a left click.
wait() #Sleep for 5s and take a screenshot to check for any changes.
finished(content='xxx') # Use escape characters \\\\', \\\\\", and \\\\n in content part to ensure we can parse the content in normal python string format. 

Coordinates are always given relative to the image size with 0 being 0% of the size and 1000 being 100%. Be as precise as possible when specifying coordinates.

## Note
- Use English in `Thought` part.
- Write a small plan and finally summarize your next action (with its target element) in one sentence in `Thought` part. 
- There is a cap of 400 characters per response.

## Level Manual
{game_instructions.GAME_INSTRUCTIONS}

## User Instruction
\end{lstlisting}

\subsubsection{User prompts}

\begin{lstlisting}[basicstyle=\normalfont, breaklines=true]
Move: Move the existing floor (caution wall) to the right so the ball does not fall off.

Extend: Stretch (extend) the existing floor at the top (roman wall) to the right so the objects to do not fall off.

Remove: Remove the wind mill from the play area.

Place: Place the balloon on the left of the play area above the top floor so that it would be blown away by the fan. No need to press the green flag.

Rotate: Flip the fan so that it blows to the right so that the wind mill starts spinning.

Multi: Place the football above the button of the flash light and move the aquarium from the left to the right to the empty space.
\end{lstlisting}

\subsection{Benchmark 5}

\subsubsection{System prompt}

\begin{lstlisting}[basicstyle=\normalfont, breaklines=true]
You are a Puzzle Game solving agent. You are given a task and your action history, with screenshots. You need to perform the next action to complete the task.

## Output Format is
```
Thought: ...
Action: ...
```
It is crucial that you stick to this output format in every single one of your responses.

## Action Space

hover(point='(x1,y1)') # Moves the mouse to the given point.
click(point='(x1,y1)') # This performs a left mouse click at the given point.
drag(start_point='(x1,y1)', end_point='(x2,y2)') # This performs a left click at the start point, moves to the end point and again performs a left click.
wait() #Sleep for 5s and take a screenshot to check for any changes.
finished(content='xxx') # Use escape characters \\\\', \\\\\", and \\\\n in content part to ensure we can parse the content in normal python string format. 

Coordinates are always given relative to the image size with 0 being 0% of the size and 1000 being 100%. Be as precise as possible when specifying coordinates.

## Note
- Use English in `Thought` part.
- Write a small plan and finally summarize your next action (with its target element) in one sentence in `Thought` part. 
- There is a cap of 400 characters per response.

## Level Manual
{game_instructions.GAME_INSTRUCTIONS}

## Full user manual
{parts_description.FULL_MANUAL}

## User Instruction
\end{lstlisting}

\subsubsection{User prompt}

\begin{lstlisting}[basicstyle=\normalfont, breaklines=true]
Solve the puzzle.
\end{lstlisting}